\documentclass[a4paper]{llncs}
\usepackage[T1]{fontenc}
\usepackage{ae}
\usepackage{aecompl}
\usepackage{microtype}
\usepackage{graphicx}
\usepackage{hyperref}
\usepackage{booktabs}
\usepackage[caption=false]{subfig}

\usepackage{amssymb}
\usepackage{amsfonts}
\usepackage{amsmath}
\usepackage{dsfont}
\usepackage{bm}

\usepackage{xcolor}

\usepackage[sectionbib,numbers,sort&compress]{natbib}

\DeclareMathOperator*{\argmin}{arg\,min}

\makeatletter
\def\blfootnote{\xdef\@thefnmark{}\@footnotetext}
\makeatother

\begin{document}

\mainmatter              
\title{Adversarial and Perceptual Refinement for Compressed Sensing MRI Reconstruction}

\author{
  Maximilian Seitzer\inst{1,2}\and
  Guang Yang\inst{3,4}\and
  Jo Schlemper\inst{2}\and
  Ozan Oktay\inst{2}\and
  Tobias W\"urfl\inst{1}\and
  Vincent Christlein\inst{1}\and
  Tom Wong\inst{3,4}\and
  Raad Mohiaddin\inst{3,4}\and
  David Firmin\inst{3,4}\and
  Jennifer Keegan\inst{3,4}\and
  Daniel Rueckert\inst{2}\and
  Andreas Maier\inst{1}
}

\institute{
  Pattern Recognition Lab, Friedrich-Alexander-Universit\"at, Erlangen, Germany\\
  \email{maximilian.seitzer@fau.de}, \and
  Biomedical Image Analysis Group, Imperial College, London, UK \and
  National Heart \& Lung Institute, Imperial College, London, UK \and
  Cardiovascular Research Centre, Royal Brompton Hospital, London, UK
}

\maketitle
\setcounter{footnote}{0} 

\begin{abstract}
Deep learning approaches have shown promising performance for compressed sensing-based Magnetic Resonance Imaging.
While deep neural networks trained with mean squared error (MSE) loss functions can achieve high peak signal to noise ratio, the reconstructed images are often blurry and lack sharp details, especially for higher undersampling rates.
Recently, adversarial and perceptual loss functions have been shown to achieve more visually appealing results.
However, it remains an open question how to (1) optimally combine these loss functions with the MSE loss function and (2) evaluate such a perceptual enhancement.
In this work, we propose a hybrid method, in which a visual refinement component is learnt on top of an MSE loss-based reconstruction network.
In addition, we introduce a semantic interpretability score, measuring the visibility of the region of interest in both ground truth and reconstructed images, which allows us to objectively quantify the usefulness of the image quality for image post-processing and analysis.
Applied on a large cardiac MRI dataset simulated with 8-fold undersampling, we demonstrate significant improvements ($p<0.01$) over the state-of-the-art in both a human observer study and the semantic interpretability score.
\end{abstract}

\section{Introduction}

\blfootnote{\noindent G. Yang and J. Schlemper/D. Rueckert and A. Maier share second/last coauthorship.}

Compressed sensing-based Magnetic Resonance Imaging (CS-MRI) is a promising paradigm allowing to accelerate MRI acquisition by reconstructing images from only a fraction of the normally required $k$-space measurements.
Traditionally, sparsity-based methods and their data-driven variants such as dictionary learning~\cite{Ravishankar11DictLearning} have been popular due to their mathematically robust formulation for perfect reconstruction.
However, these methods are limited in acceleration factor and also suffer from high computational complexity.
More recently, several deep learning-based architectures have been proposed as an attractive alternative for CS-MRI.
The advantages of these techniques are their computational efficiency, which enables real-time application, and that they can learn powerful priors directly from the data, which allows higher acceleration rates.
The most widely adopted deep learning approach is to perform an end-to-end reconstruction using multi-scale encoding-decoding architectures~\cite{Lee17ResidualMRI, Yang18DAGAN}.
Alternative approaches carry out the reconstruction in an iterative manner~\cite{Yang16DeepADMM}, conceptually extending traditional optimization algorithms.
Most previous studies focus on exploring the network architecture; however, the optimal loss function to train the network remains an open question.

Recently, as an alternative to the commonly used MSE loss, adversarial~\cite{Dosovitskiy16PercSim} and perceptual losses~\cite{Johnson16PerceptualLosses} have been proposed for CS-MRI~\cite{Yang18DAGAN}.
As these loss functions are designed to improve the visual quality of the reconstructed images, we refer to them as \emph{visual loss functions} in the following.
So far, approaches using visual loss functions still rely on an additional MSE loss for successful training of the network.
Directly combining all these losses in a joint optimization leads to a suboptimal training process resulting in reconstructions with lower peak signal to noise ratio (PSNR) values.
In this work, we propose a two-stage architecture that avoids this problem by separating the reconstruction task from the task of refining the visual quality.
Our contributions are the following:
(1) we show that the proposed refinement architecture improves visual quality of reconstructions without compromising PSNR much, and
(2) we introduce the semantic interpretability score as a new metric to evaluate reconstruction performance, and show that our approach outperforms competing methods on it.

\section{Background}
\noindent\textbf{Deep Learning-based CS-MRI Reconstruction.~}
Let $x \in \mathds{C}^N$ denote a complex-valued MR image of size $N$ to be reconstructed, and let $y \in \mathds{C}^M$ $(M << N)$ represent undersampled $k$-space measurements obtained by $y = F_u x + \varepsilon$, where $F_u$ is the undersampling Fourier encoding operator and $\varepsilon$ is complex Gaussian noise.
The linear inversion $x_u = F_u^H y$, also called zero-filled reconstruction, is fundamentally ill-posed and generates an aliased image due to violation of the Nyquist-Shannon sampling theorem.
Therefore, it is necessary to add prior knowledge into the reconstruction to constrain the solution space, traditionally formulated as the following optimization problem:
\begin{equation}
  \setlength{\abovedisplayskip}{0.5em}
  \setlength{\belowdisplayskip}{0.25em}
  \label{eq:sparse_coding}
\argmin_x \; \mathcal{R}(x) + \lambda \lVert y - F_u x \rVert^2_2
\end{equation}
Here, $\mathcal{R}$ expresses a regularization term on $x$ (e.\,g.\ $\ell_0/\ell_1$-norm for CS-MRI), and $\lambda$ is a hyper-parameter reflecting the noise level. In deep learning
approaches, one learns the inversion mapping directly from the data.
However, rather than learning the mapping from Fourier directly to image domain, it is common to formulate this problem as \emph{de-aliasing} the zero-filled reconstructions $x_u$ in the image domain~\cite{Schlemper17RecNet,Yang18DAGAN}.
Let $\mathcal{D}$ be our training dataset of pairs $(x, x_u)$ and $\hat{x} = R(x_u)$ be the image generated by the reconstruction network $R$.
Given $\mathcal{D}$, the network is trained by minimizing the empirical risk $\mathcal{L}(R) = \mathbb{E}_{(x, x_u) \sim \mathcal{D}} \, d(x, \hat{x})$, where $d$ is a distance function measuring the dissimilarities between the reference fully-sampled image and the reconstruction.

For the choice of the reconstruction network $R$, most previous approaches~\cite{Lee17ResidualMRI,Yang18DAGAN} relied on an encoder-decoder structure (e.\,g.\ U-Net~\cite{Ronneberger15UNET}), but our preliminary experiments showed that these architectures performed subpar in terms of PSNR.
Instead, we use the architecture proposed in~\cite{Schlemper17RecNet}, as it performed well even for high undersampling rates.
This network consists of $n_c$ consecutive de-aliasing blocks, each containing $n_d$ convolutional layers.
Each de-aliasing block takes an aliased image $x^{(i)} \in \mathbb{R}^{2N}$ as the input and outputs the de-aliased image $x^{(i+1)} \in \mathbb{R}^{2N}$, with $i \in \{0, \dots n_c-1\}$ and $x^{0} = x_u = F^H_u y$ being the zero-filled reconstruction.
Interleaved between the de-aliasing blocks are data consistency (DC) layers, which enforce that the reconstruction is consistent with the acquired $k$-space measurements by replacing frequencies of the intermediate image with frequencies retained from the sampling process.
This process can be seen as an unrolled iterative reconstruction where de-aliasing blocks and DC layers perform the role of the regularization step and data fidelity step, respectively~\cite{Schlemper17RecNet}.\\

\noindent\textbf{Loss Functions for Reconstruction.~}
In deep learning-based approaches to inverse problems, such as MR reconstruction and single image super-resolution, a frequently used loss function~\cite{Lee17ResidualMRI, Yang16DeepADMM} is the MSE loss $\mathcal{L}_{\text{MSE}}(R) = \mathbb{E}_{(x, x_u) \sim \mathcal{D}} {\lVert x - \hat{x} \rVert}^2_2$.
Though networks trained with MSE criterion can achieve high PSNR, the results often lack high frequency image details~\cite{Dahl17PixelSR}.
Perceptual loss functions~\cite{Johnson16PerceptualLosses} are an alternative to the MSE loss.
They minimize the distance to the target image in some feature space.
A common perceptual loss is the VGG loss $\mathcal{L}_{\text{VGG}}(R) = \mathbb{E}_{(x, x_u) \sim \mathcal{D}} {\lVert f_{\text{VGG}}(x) - f_{\text{VGG}}(\hat{x})\rVert}^2_2$, where $f_{\text{VGG}}$ denotes VGG feature maps~\cite{Simonyan15VGG}.

Another choice is an adversarial loss based on Generative Adversarial Networks (GANs)~\cite{Goodfellow14GAN, Dosovitskiy16PercSim}.
A discriminator and a generator network are setup to compete against each other such that the discriminator is trained to differentiate between real and generated samples, whereas the generator is encouraged to deceive the discriminator by producing more realistic samples.
For us, the discriminator $D$ learns to differentiate between fully-sampled and reconstructed images, and the reconstruction network, playing the role of the generator, reacts by changing the reconstructions to be more similar to the fully-sampled images.
The discriminator loss is then given by $\mathcal{L}_{\text{GAN}}(D) = -\mathbb{E}_{(x, x_u) \sim \mathcal{D}} \log(D(x)) + \log(1-D(R(x_u)))$.
During training, the reconstruction network minimizes $\mathcal{L}_{\text{adv}}(R) = -\mathbb{E}_{(x, x_u) \sim \mathcal{D}} \log(D(R(x_u)))$, which has the effect of pulling the reconstructed images closer towards the distribution of the training data.

Perceptual losses are known to increase textural details~\cite{Ledig17SR}, but also to introduce high frequency artifacts~\cite{Dosovitskiy16PercSim}, whereas adversarial losses can produce realistic, high frequency details~\cite{Ledig17SR}.
As perceptual and adversarial losses complement each other, it is sensible to combine them into a single visual loss $L_{\text{vis}}(R) = \mathcal{L}_{\text{adv}}(R) + \mathcal{L}_{\text{VGG}}(R)$.
For MR reconstruction, previous attempts~\cite{Yang18DAGAN} further combined adversarial and/or perceptual loss with the MSE loss to stabilize the training.
This simultaneous optimization yields acceptable solutions, typically however with low PSNR.
We argue this is because the different training objectives compete with each other, leading to the network ultimately converging to a suboptimal local maximum.

\section{Method}

\begin{figure}[t]
  \centering
  \subfloat[Stage 1: training of reconstruction network using $\mathcal{L}_{\text{MSE}}(R)$.]{
    \makebox[\textwidth]{\includegraphics[width=0.75\textwidth]{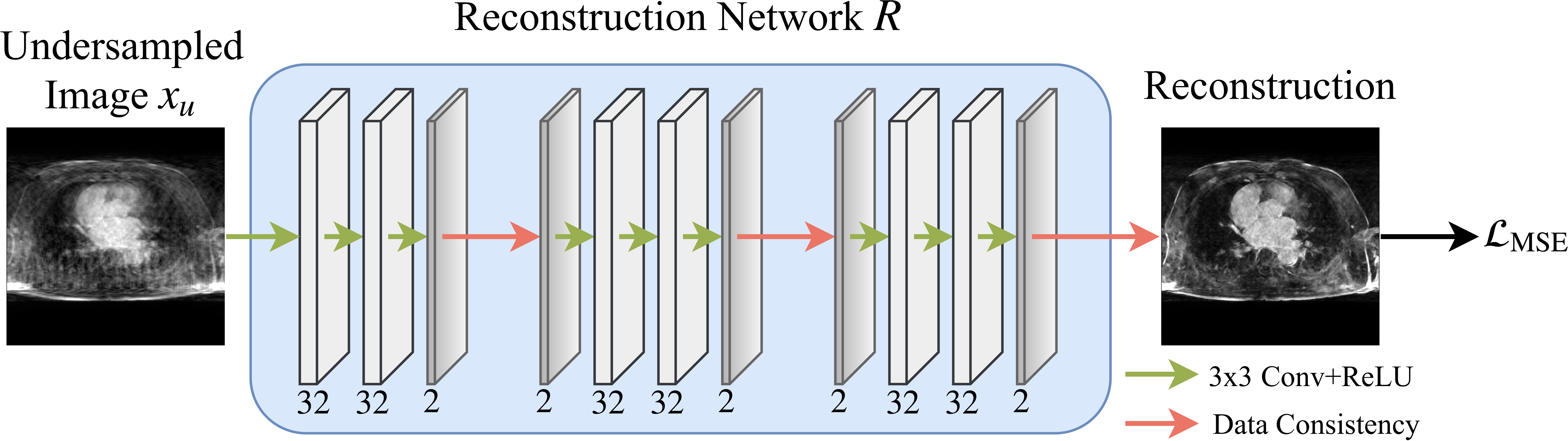}}
  }

  \subfloat[Stage 2: training of visual refinement network using $\mathcal{L}_{\text{vis}}(V)$.]{
    \makebox[\textwidth]{\includegraphics[width=\textwidth]{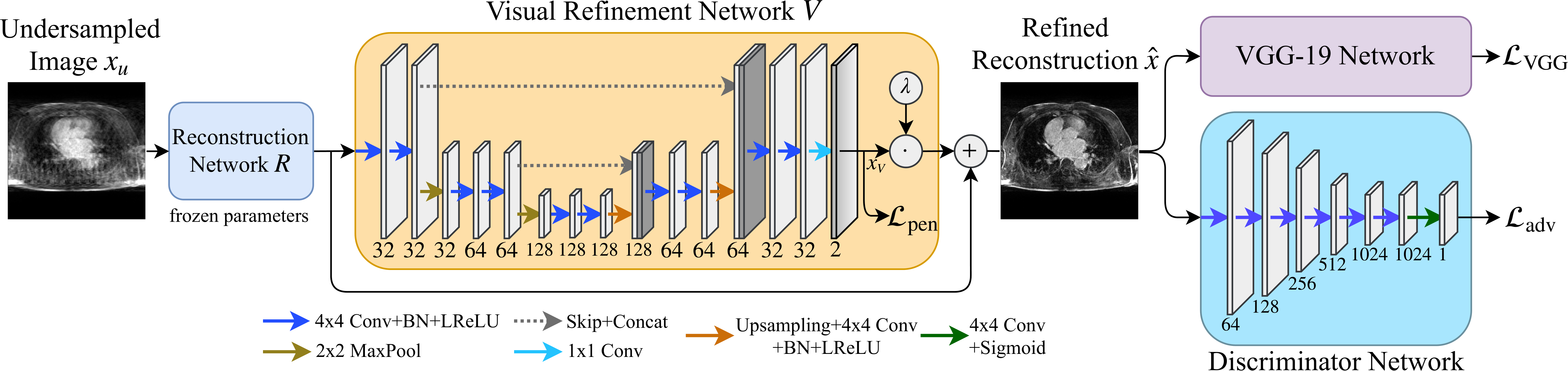}}
  }
  \caption{Overview of proposed method.}
  \label{fig:architecture}
\end{figure}

The observation above motivates our approach:
instead of directly training a reconstruction network with all loss functions jointly, we use a two-stage procedure, detailed in Figure \ref{fig:architecture}.
In the first stage, the reconstruction network $R$ is trained with $\mathcal{L}_{\text{MSE}}(R)$.
In the second stage, we fix the reconstruction network and train a visual refinement network $V$ on top of $R$ by optimizing $\mathcal{L}_{\text{vis}}(V)$.
The final reconstruction is then given by $\hat{x} = R(x_u) + V(R(x_u))$, i.\,e.\, $V$ learns an additive mapping which refines the base reconstruction.
In this setup, discriminator and VGG network still receive the full reconstruction $\hat{x}$ as input.

The decoupling of the refinement step from the reconstruction task has several benefits.
The discriminator begins training by seeing reasonably good reconstructions, which avoids overfitting it to suboptimal solutions during the training process.
Furthermore, compared to training from scratch, the optimization is easier as it starts closer to the global optimum.
Finally, the visual refinement step always starts out from the best possible MSE solution achievable with $R$, whereas this guarantee is not given when jointly training $R$ with $\mathcal{L}_{\text{MSE}}$ and $\mathcal{L}_{\text{vis}}$.

The choice of the architecture for the visual refinement network is flexible, and in this work we use a U-Net architecture.
Within $V$, we gate the output of the network by a trainable scalar $\lambda$, which improves the adversarial training dynamics during the early stages of training.
If we initialize $\lambda = 0$, the discriminator receives $\hat{x} = R(x_u)$, and the gradient signal to $V$ is forced to zero.
This allows the discriminator to initially only learn on clean reconstructions from $R$, untainted by the randomly initialized output of $V$.
For the refinement network, the impact of less useful gradients is reduced while the discriminator has not yet learned to correctly differentiate between the ground truth (i.\,e.\, fully-sampled data) and the reconstructions.
We also scale $R(x_u)$ to the range of $(-1, 1)$ before using it as $V$'s input and then scale $\hat{x}$ back to the original range after adding the refinement.
In accordance to our goal of reaching high PSNR values, we constrain the output $x_{\scriptscriptstyle V}$ of $V$ (before gating with $\lambda$) with an $\ell^1$-penalty $\mathcal{L}_{\text{pen}}(V) = \lVert x_{\scriptscriptstyle V}\rVert_1$.
This guides $V$ to learn the minimal sparse transformation needed to fulfill the visual loss, i.\,e.\, to change the MSE-optimal solution only in areas important for visual quality.
In practice, this means that our approach yields higher PSNR values compared to joint training, as we show in section~\ref{sec:experiments}.

We also utilize a couple of techniques known to stabilize the adversarial learning process.
For the discriminator network, we use one-sided label smoothing~\cite{Salimans16ImprovedGAN} of $0.1$, and an experience replay buffer~\cite{Pfau16GAN} of size $80$ with probability $p=0.5$ to draw from it.
For the refinement network, we add a feature matching loss~\cite{Salimans16ImprovedGAN} $\mathcal{L}_{\text{feat}}(V) = \mathbb{E}_{(x, x_u) \sim \mathcal{D}} \frac{1}{N} \sum_{i=1}^{N} \lVert f_{D}^{(i)}(x) - f_{D}^{(i)}(\hat{x}) \rVert_1$, where $f_{D}^{(i)}$ denotes the $i$'th of $N$ feature maps of the discriminator.
The total loss for $V$ is given by
\begin{equation}
  \setlength{\abovedisplayskip}{0.5em}
  \setlength{\belowdisplayskip}{0.5em}
  \label{eq:total_loss}
  \begin{aligned}
    \mathcal{L}(V) = \frac{1}{2} \bigg(\frac{\mathcal{L}_{\text{adv}}(V)}{M}  + \frac{\mathcal{L}_{\text{feat}}(V)}{N} \bigg) + \frac{\mathcal{L}_{\text{VGG}}(V)}{O} + \alpha \mathcal{L}_{\text{pen}}(V)
  \end{aligned}
\end{equation} with $\alpha$ being the penalty strength, and $M$, $N$, $O$ constants set such that $\frac{\mathcal{L}_{\text{adv}}}{M} = \frac{\mathcal{L}_{\text{feat}}}{N} = \frac{\mathcal{L}_{VGG}}{O} = 1$ in the first iteration of training, which amounts to assigning the two adversarial loss terms the same initial importance as $\mathcal{L}_{\text{VGG}}$.
The penalty strength $\alpha$ is important for training speed and stability.
Choosing $\alpha$ such that $\mathcal{L}_{\text{pen}} \approx 0.1$ in the first training iteration gave us sufficiently good results.\\

\noindent\textbf{Semantic interpretability score.~}
The most commonly used metrics to evaluate reconstruction quality are PSNR and the structural similarity index (SSIM). 
It has been shown that those two metrics do not necessarily correspond to visual quality for human observers, as e.\,g.\ demonstrated by human observer studies in~\cite{Ledig17SR, Dahl17PixelSR}.
Therefore, PSNR and SSIM alone are not sufficient in the evaluation of image reconstructions.
This poses the question on how to evaluate reconstruction quality taking human perception into account.
One possibility is to let domain experts (e.\,g.\, clinicians and MRI physicists) rate the reconstructions and average the results to form a mean opinion score (MOS).
Obtaining opinion scores from expert observers is costly, hence cannot be used during the development of new models.
However, if expert-provided segmentation labels are available, we can design a metric indicating how visible the segmented objects are in the reconstructed images, in the following referred to as \emph{semantic interpretability score} (SIS).
This metric is motivated by Inception scores~\cite{Salimans16ImprovedGAN} in GANs, which tells how well an Inception network can identify objects in generated images.

SIS is defined as the mean Dice overlap between the ground truth segmentation and the segmentation predicted by a pre-trained segmentation network from the reconstructed images.
The scores are normalized by the average Dice score on the ground-truth images to obtain a measure of segmentation performance relative to the lower error-bound.
We only consider images in which at least one instance of the object class is present, and ignore the background class.
We argue that if a pre-trained network is able to produce better segmentations, the regions of interest are better visible (e.\,g.\ have clearly defined boundaries) in the images.
Implementing SIS requires a segmentation network trained on the same distribution of images as the reconstruction dataset.
In practice, the segmentation network is trained on the fully-sampled images used for training the reconstruction method.
We trained an off-the-shelf U-Net architecture to segment the left atrium, achieving a Dice score of 0.796 on the ground truth images.

\section{Experiments}
\label{sec:experiments}
\noindent\textbf{Datasets.~}
We evaluated our method on 3D late gadolinium enhanced cardiac MRI datasets acquired in 37 patients.
We split the 2D axial slices of the 3D volumes into 1248 training images, 312 validation images, and 364 testing images of size $512\times512$ pixels.
For training, we generated random 1D-Gaussian masks keeping 12.5\% of raw $k$-space data, which corresponded to an 8$\times$ speed-up.
During testing, we randomly generated a mask for each slice, which we kept the same for all evaluated methods.\\

\noindent\textbf{Training Details and Parameters.~}For the reconstruction network, we used $n_c=3$ de-aliasing blocks, and $n_d=3$ convolutional layers with 32 filters of size 3$\times$3.
For the refinement network, we used a U-Net with 32, 64, 128 encoding filters and 64, 32 decoding filters of size 4x4, batch normalization and leaky ReLU with slope 0.1.
The discriminator used a PatchGAN~\cite{Isola17ImagetoImage} architecture with 64, 128, 256, 512, 1024, 1024 filters of size 4x4, and channelwise dropout after the last 3 layers.
The VGG loss used the final convolutional feature maps of a VGG-19 network pre-trained on ImageNet.
The reconstruction network was trained for 1500 epochs with batch size 20, the refinement network for 200 epochs with batch size 5, both using the Adam optimizer~\cite{Kingma15Adam} with learning rate 0.0002, $\beta_1=0.5$, $\beta_2=0.999$.
We found that the training is sensitive to the network's initialization.
Thus, we chose orthogonal initialization~\cite{Saxe14OrthInit} for the refinement network and Gaussian initialization from $\mathcal{N}(0, 0.02)$ for the discriminator.\\

\noindent\textbf{Evaluation Metrics.~}
We use PSNR and SIS as evaluation metrics.
To further evaluate our approach and assess how useful SIS is as a proxy for visual quality, we also asked a domain expert to rate all reconstructed images in which the left atrium anatomy and the atrial scars are visible.
The rating ranges from 1 (poor) to 4 (very good), and is based on the overall image quality, the visibility of the atrial scar and occurrence of artifacts.
To obtain an unbiased rating, the expert was shown all images from all methods in randomized order.\\

\noindent\textbf{Results.~}We compared our approach against three other reconstruction methods:
RecNet\footnote{\url{https://github.com/js3611/Deep-MRI-Reconstruction}}~\cite{Schlemper17RecNet} (i.e. the proposed approach without refinement step), DAGAN\footnote{\url{https://github.com/nebulaV/DAGAN}}~\cite{Yang18DAGAN} using both adversarial and perceptual loss, and DLMRI\footnote{\url{http://www.ifp.illinois.edu/~yoram/DLMRI-Lab/DLMRI.html}}~\cite{Ravishankar11DictLearning}, a dictionary learning based method. No data augmentation was used for any of the methods.

{
\setlength{\belowcaptionskip}{0em}
\setlength{\textfloatsep}{0em}
\begin{table}[tb!]
  \begin{center}
  \scalebox{.9}{
{\renewcommand{\arraystretch}{1.2}\setlength{\tabcolsep}{1em}
  \begin{tabular}[c]{ccccc}
  \toprule
  Method                                 & PSNR (dB)                  & MOS                      & SIS        \\
  \midrule
   Ground Truth                           & $\infty$                  & $3.78 \pm 0.45$          & $1$             \\
   \midrule
  RecNet \cite{Schlemper17RecNet}        & $\mathbf{32.46} \pm 2.26$  & $2.75 \pm 0.78$          & $0.801$          \\
  DLMRI \cite{Ravishankar11DictLearning} & $31.45 \pm 2.40$           & $1.09 \pm 0.29$          & $0.842$          \\
  DAGAN \cite{Yang18DAGAN}               & $28.41 \pm 1.91$           & $2.61 \pm 0.83$          & $0.812$          \\
    Proposed Method                        & $31.89 \pm 2.18$         & $\mathbf{3.24} \pm 0.63$ & $\mathbf{0.941}$ \\
  \bottomrule
  \end{tabular}}
  }
  \end{center}
  \caption{Quantitative results for 8-fold undersampling. Highest measures in bold.}
  \label{table:results}
\end{table}}

\begin{figure}[tb]
  \centering
  \setlength{\belowcaptionskip}{-0.25em}
  \setlength{\belowcaptionskip}{-1em}
  \subfloat[Zero-filled]{
    \includegraphics[width=0.28\textwidth,trim={0.28cm 0.28cm 0.2cm 0.2cm},clip,angle=90]{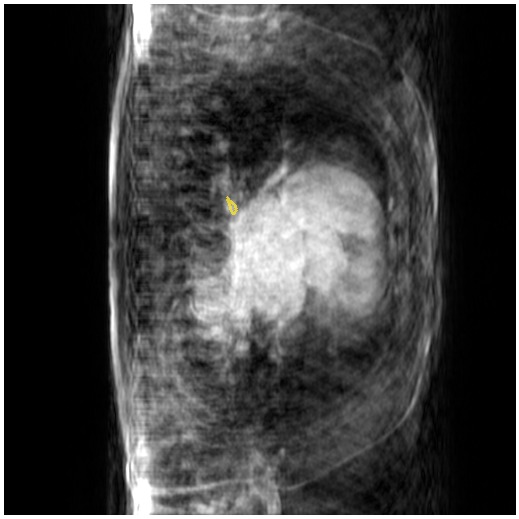}
  }\quad
  \subfloat[DLMRI]{
    \includegraphics[width=0.28\textwidth,trim={0.28cm 0.28cm 0.2cm 0.2cm},clip,angle=90]{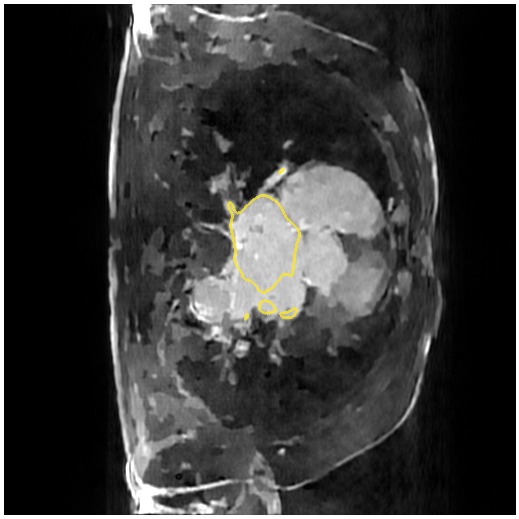}
  }\quad
  \subfloat[DAGAN]{
    \includegraphics[width=0.28\textwidth,trim={0.28cm 0.28cm 0.2cm 0.2cm},clip,angle=90]{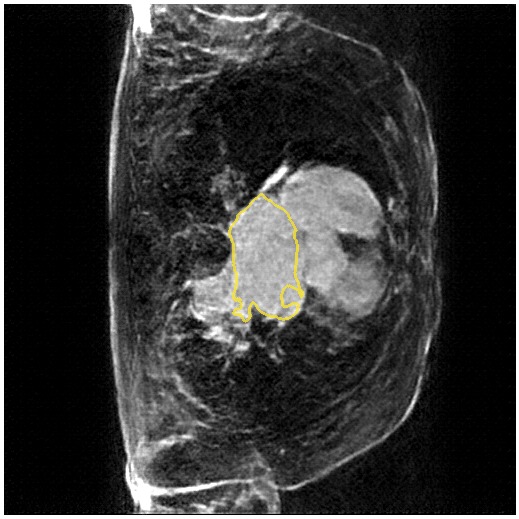}
  }
  \vspace{-0.8em}
  \subfloat[RecNet]{
    \includegraphics[width=0.28\textwidth,trim={0.28cm 0.28cm 0.2cm 0.2cm},clip,angle=90]{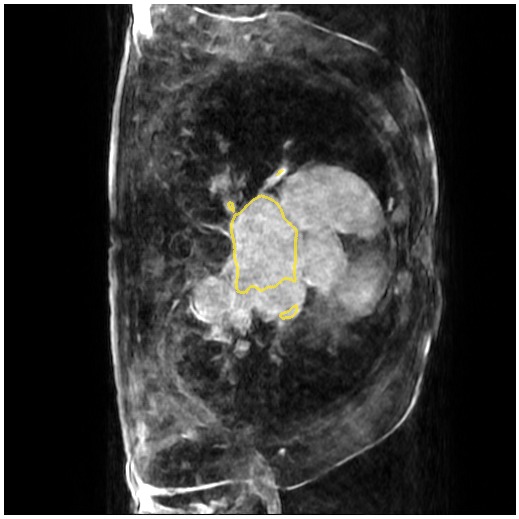}
  }\quad
  \subfloat[Proposed Method]{
    \includegraphics[width=0.28\textwidth,trim={0.28cm 0.28cm 0.2cm 0.2cm},clip,angle=90]{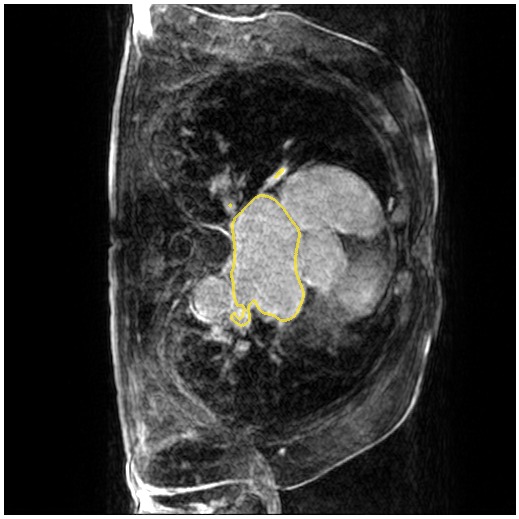}
  }\quad
  \subfloat[Ground Truth]{
    \includegraphics[width=0.28\textwidth,trim={0.28cm 0.28cm 0.2cm 0.2cm},clip,angle=90]{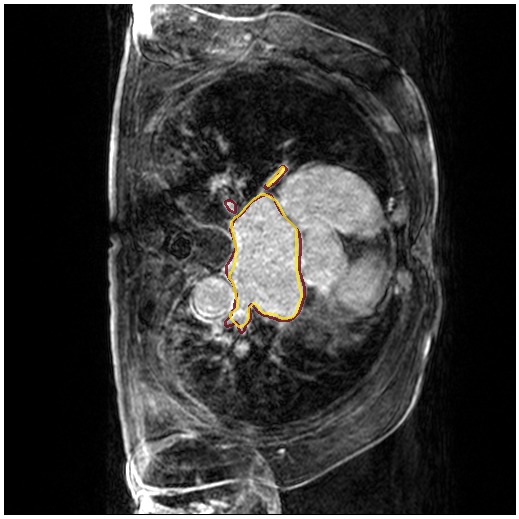}
    \label{subfig:sample_gt}
  }
  \caption{Qualitative visualization for 8-fold undersampling. Contour of predicted segmentation of left atrium in yellow, contour of ground truth segmentation in red.}
  \label{fig:samples}
\end{figure}

We show the results of our evaluation in Table~\ref{table:results}, and a sample reconstruction in Figure~\ref{fig:samples}.
RecNet performed best in terms of PSNR, which is expected as its training objective directly corresponds to this metric, but its reconstructions were over-smoothed.
DLMRI had the lowest MOS, with its reconstructions showing heavy oil paint artifacts.
DAGAN, combining MSE loss with a visual loss function without any further precautions, suffered from low PSNR.
While its reconstructions also looked sharp, they were noisy and often displayed aliasing artifacts, which was reflected in a lower MOS compared to our method.
Our proposed approach achieved significantly\footnote{Significance determined by a two-sided paired Wilcoxon signed-rank test at $p < 0.01$.} higher mean opinion score than all other methods, while still maintaining high PSNR.
Reconstructions obtained by our method appeared sharper with better contrast.
Moreover, our method achieved the highest SIS close to segmentation performance on the ground truth data, which indicated that the segmented objects were clearly visible in the reconstructed images.\looseness=-1

These results further demonstrate that PSNR alone is a subpar indicator for reconstruction quality, making our SIS a useful supplement to those metrics.
For our method, SIS agreed with the quality score given by the expert user.
Somewhat surprising is that the SIS of DLMRI is slightly higher than RecNet and DAGAN although DLMRI has the worst MOS.
We conjecture this is because, although DLMRI reconstructed images lack textural details, areas belonging to the same organ have similar intensity values, which helps the segmentation task.
While scoring through an expert user is thus still the safest way to evaluate reconstructions, we believe that in conjunction with PSNR, SIS is a helpful tool to quickly judge image quality during the development of new models.

\section{Conclusion}

In this work, we highlighted the inadequacy of previously proposed deep learning based CS-MRI methods using MSE loss functions in direct combination with visual loss functions.
We improved on them by proposing a new refinement approach, which incorporates both loss functions in a harmonious way to improve the training stability.
We demonstrated that our method can produce high quality reconstructions with large undersampling factors, while keeping higher PSNR values compared to other state-of-the-art methods.
We also showed that the reconstruction obtained by our method can provide the best segmentation of the ROIs among all compared methods.

{\small
 \bibliographystyle{splncs03}
 \bibliography{paper}
}

\appendix
\section{Appendix}

The following images show more samples for 8-fold undersampling.
For each of the seven patients of the test set, a random slice showing the left atrium was selected. 
The contour of the predicted segmentation of left atrium is shown in yellow, the contour of the ground truth segmentation in red.

\noindent
\includegraphics[width=\textwidth]{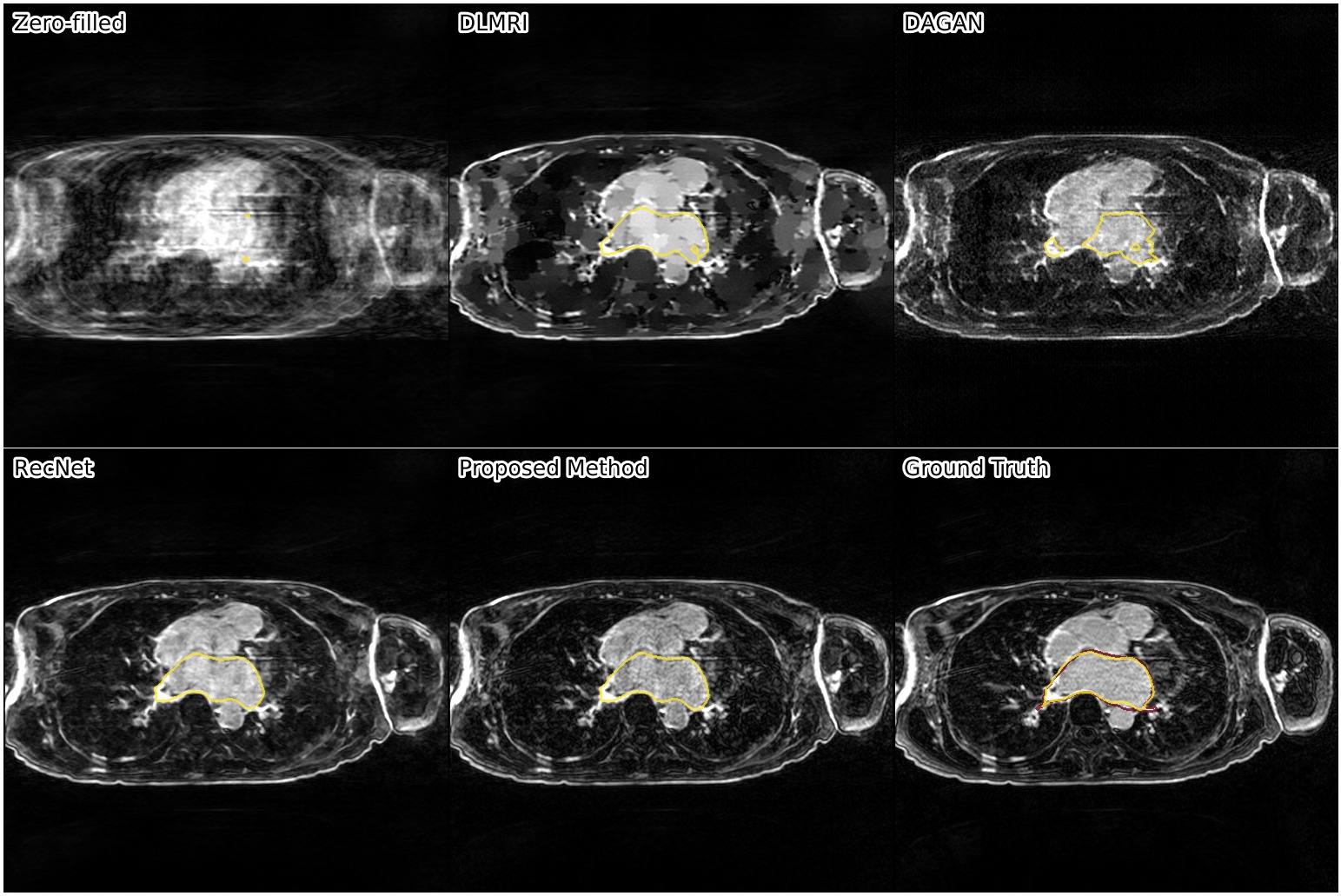}

\noindent
\includegraphics[width=\textwidth]{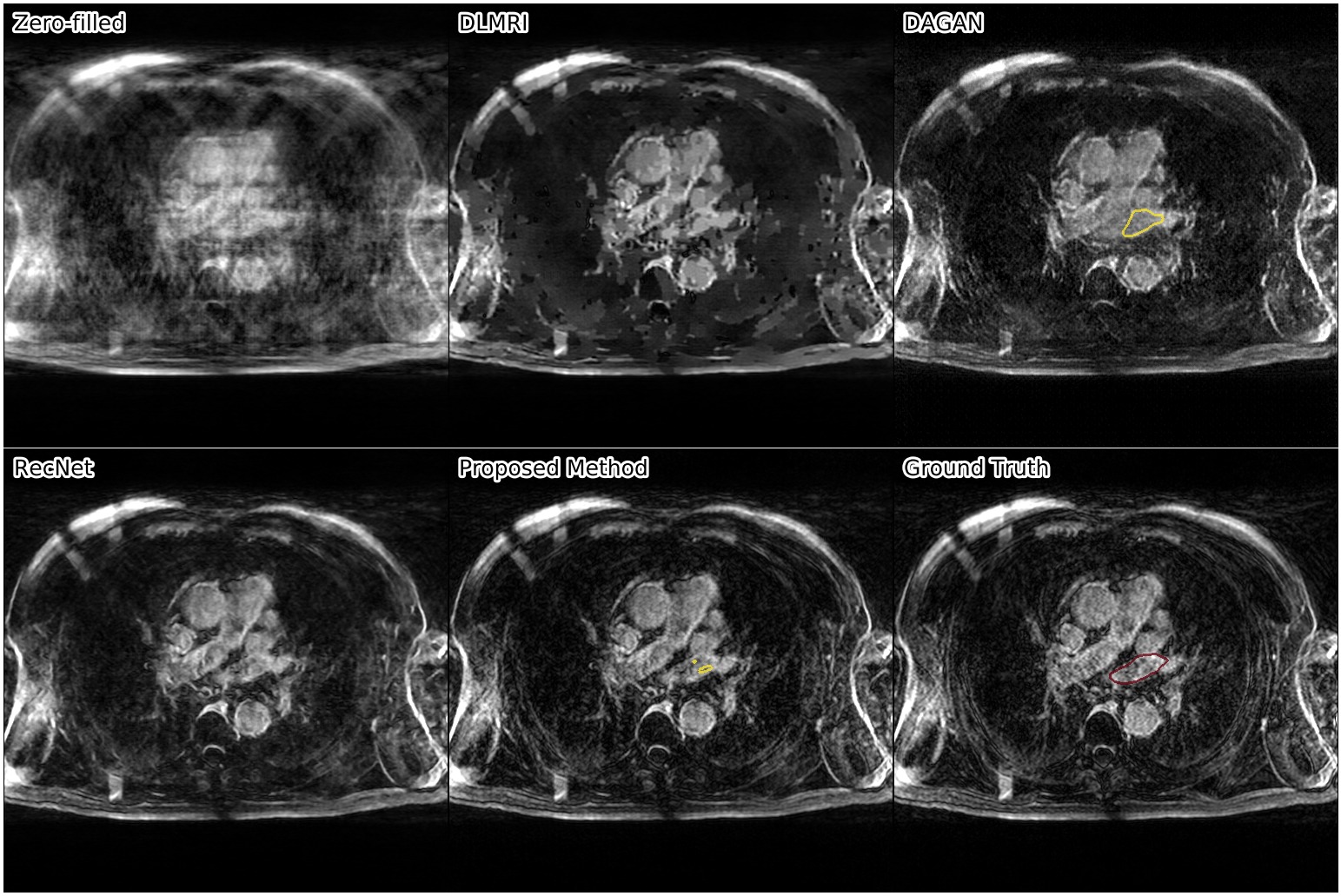}
\includegraphics[width=\textwidth]{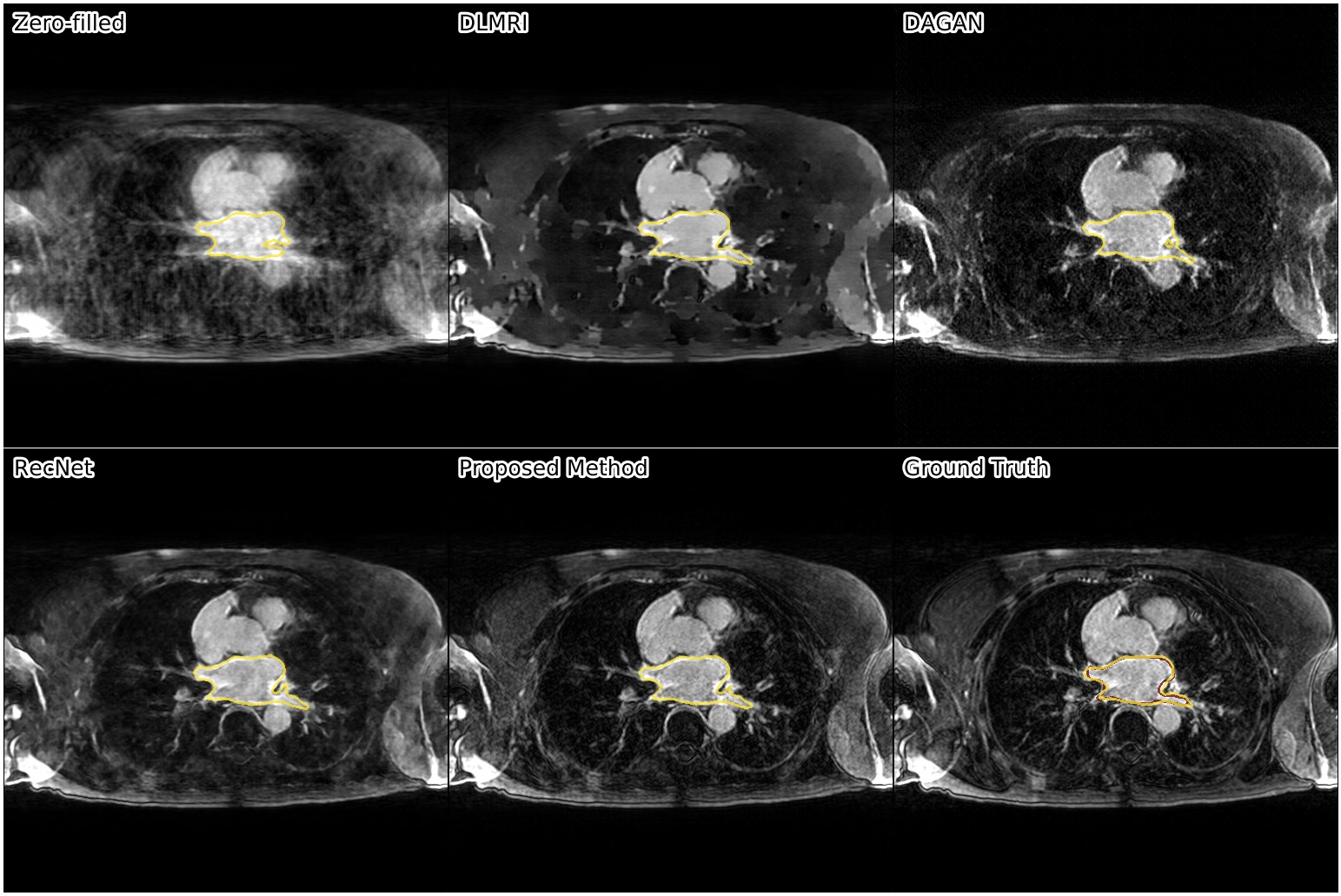}
\includegraphics[width=\textwidth]{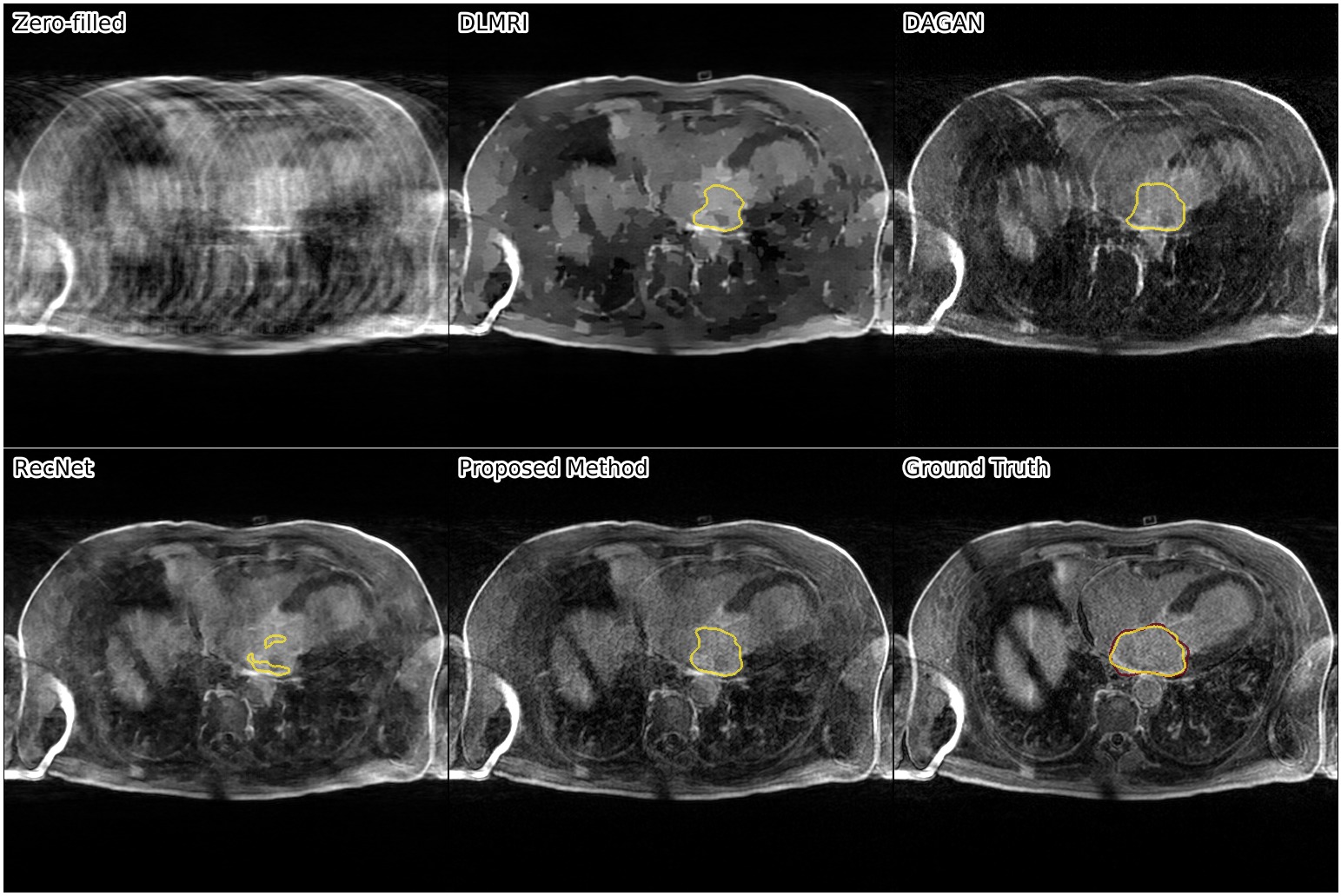}
\includegraphics[width=\textwidth]{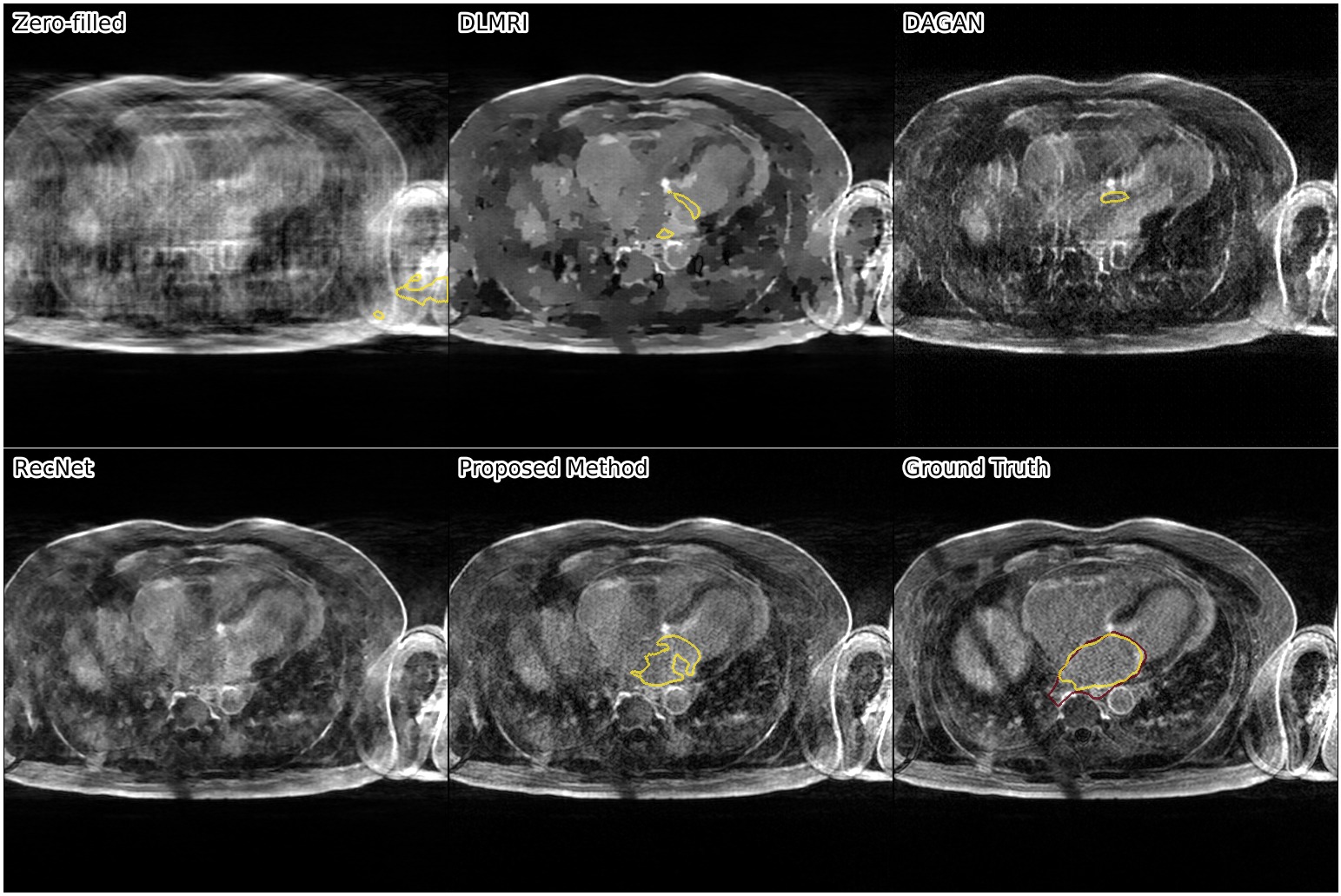}
\includegraphics[width=\textwidth]{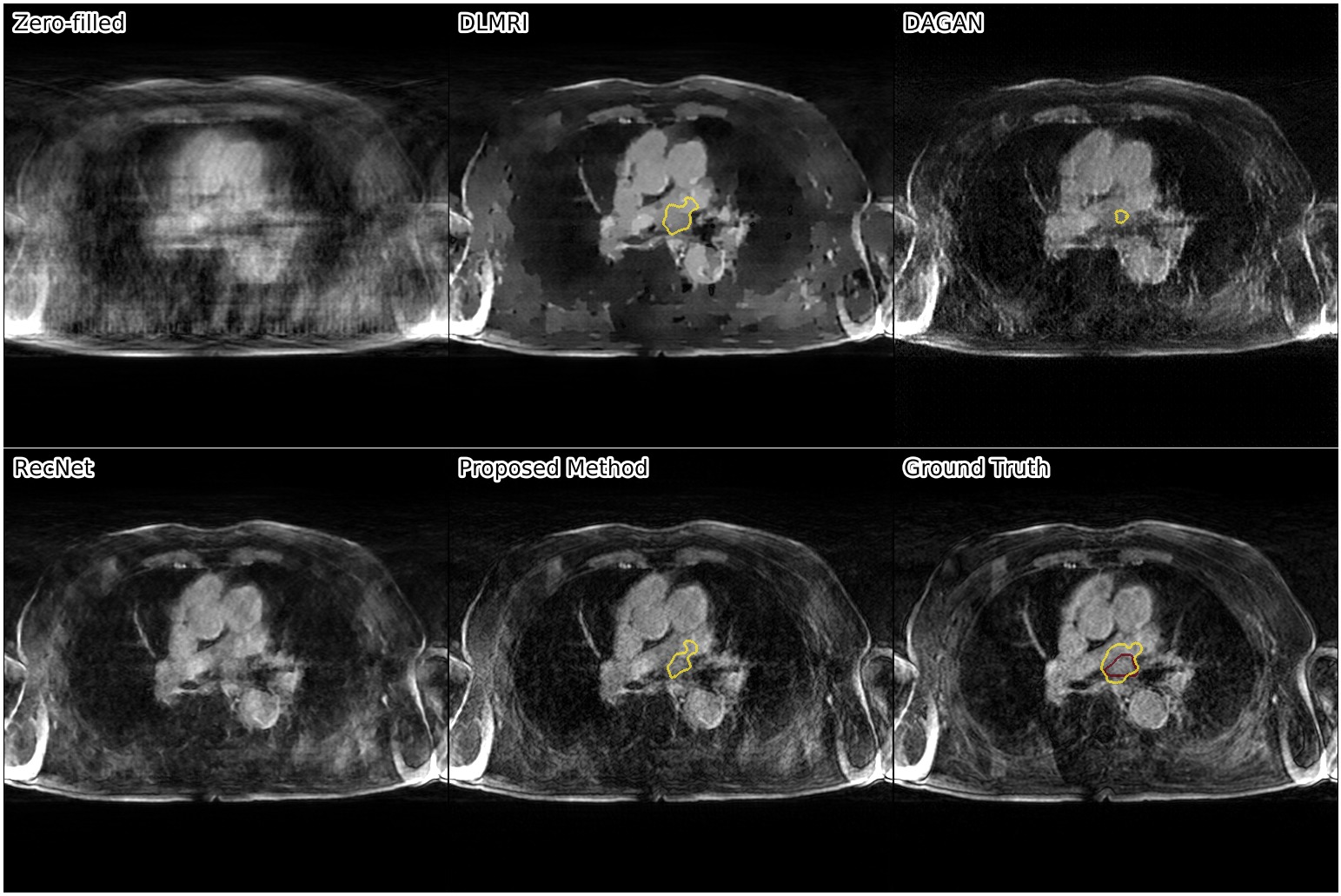}
\includegraphics[width=\textwidth]{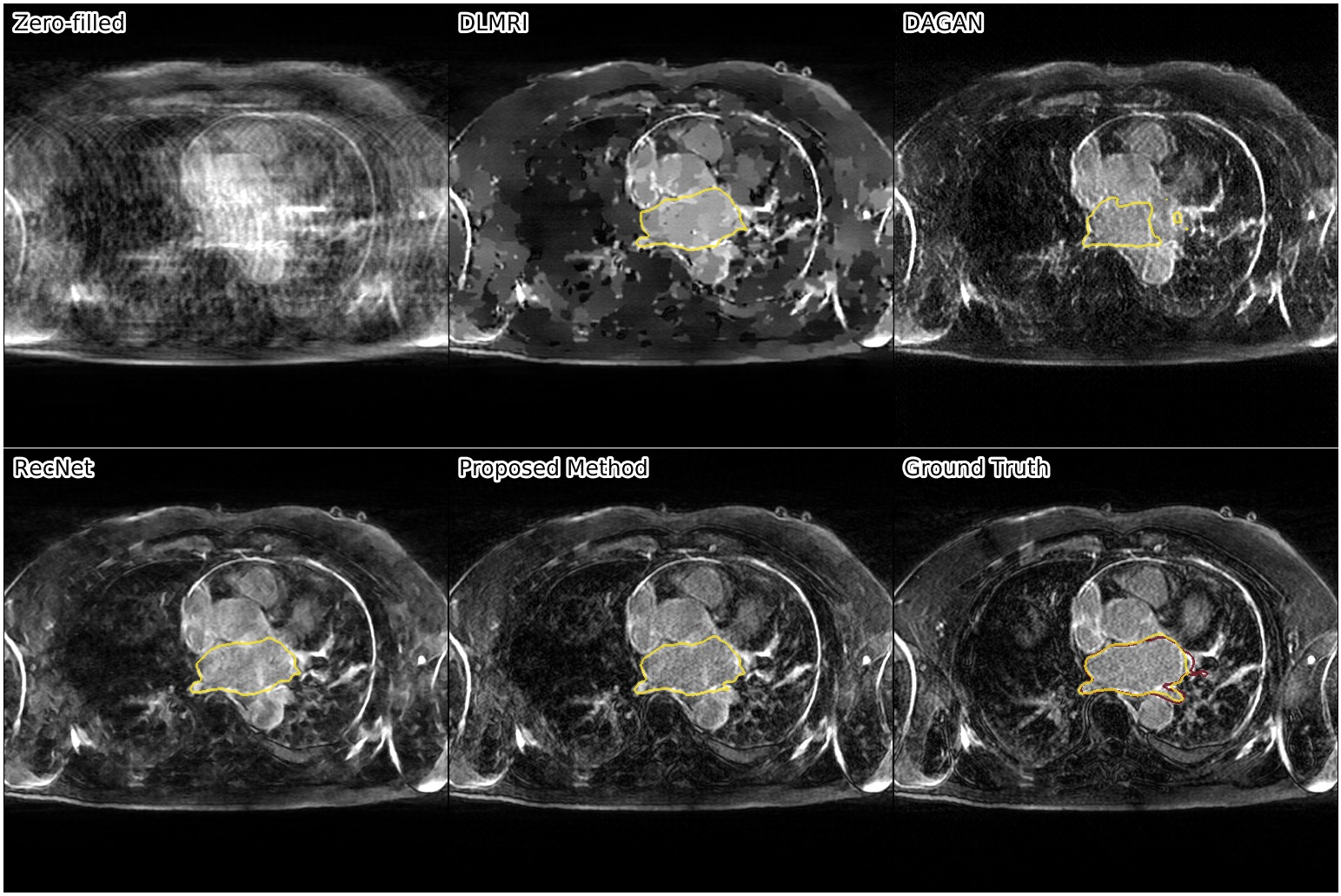}

\end{document}